\documentclass[a4paper, 10pt, conference]{ieeeconf}

\IEEEoverridecommandlockouts                              
\usepackage{graphicx}
\usepackage{url}
\usepackage{pifont}
\usepackage{color}
\usepackage{amsmath}
\usepackage{tabularx} 
\usepackage{algorithm}
\usepackage{algorithmic}
\usepackage{amsfonts}
\usepackage{multirow} 
\usepackage{multicol} 

\title{\LARGE \bf
RMAU-NET: A Residual-Multihead-Attention U-Net Architecture for Landslide Segmentation and Detection from Remote Sensing Images 
}

\author{Lam~Pham$^{*}$,
        Cam~Le$^{*}$, 
        Hieu~Tang$^{*}$, 
        Khang~Truong$^{*}$,
        Truong~Nguyen,
        Jasmin~Lampert, \\        
        Alexander~Schindler,
        Martin~Boyer,
        Son~Phan$^{\dagger}$      
\thanks{L. Pham is with Van Lang University, Vietnam.}
\thanks{L. Pham, C. Le, J. Lampert, A. Schindler, and M. Boyer are with the Competence Unit Data Science \& Artificial Intelligence at the Austrian Institute of Technology, Austria.}%
\thanks{H. Tang is with the Troyes University, France.}%
\thanks{K. Truong and T. Nguyen are with HCM University of Technology, Vietnam.}%
\thanks{S. Phan is with Ton Duc Thang University, Vietnam.}%
\thanks{(*) Main and equal contribution into the paper.}
\thanks{($\dagger$) Corresponding author.}
}

\begin{document}

\maketitle
\thispagestyle{empty}
\pagestyle{empty}

\begin{abstract}
In recent years, landslide disasters have reported frequently due to the extreme weather events of droughts, floods , storms,  or the consequence of human activities such as deforestation, excessive exploitation of natural resources. 
However, automatically observing landslide is challenging due to the extremely large observing area and the rugged topography such as mountain or highland.
This motivates us to propose an end-to-end deep-learning-based model which explores the remote sensing images for automatically observing  landslide events.
By considering remote sensing images as the input data, we can obtain free resource, observe large and rough terrains by time.
To explore the remote sensing images, we proposed a novel neural network architecture which is for two tasks of landslide detection and landslide segmentation.
We evaluated our proposed model on three different benchmark datasets of LandSlide4Sense, Bijie, and Nepal.
By conducting extensive experiments, we achieve F1 scores of 98.23, 93.83 for the landslide detection task on LandSlide4Sense, Bijie datasets; mIoU scores of 63.74, 76.88 on the segmentation tasks regarding LandSlide4Sense, Nepal datasets.
These experimental results prove potential to integrate our proposed model into real-life landslide observation systems. 

%

\indent \textit{Items}--- Attention, landslide detection, landslide segmentation, remote sensing image, U-Net.
\end{abstract}


\section{INTRODUCTION}
\label{intro}
According to the literature, landslides often occur due to the instability of slopes~\cite{disaster_charter_org, picarelli2021impact, emdat}. 
During such events, soil, rock, mud, or debris from geological activities collapse and slide out/down from hills or mountains, causing significant damage across multiple aspects of human life, including loss of life, psychological trauma after suffering the event, demolishing of agriculture, and long-term impacts on communities near the affected sites~\cite{doi:https://doi.org/10.1002/9780470012659.ch2}. 
For example, a landslide in the Indian state of Kerala in July 2024 resulted in 24 fatalities~\cite{Landslide_in_India}, while another landslide in Ethiopia during the same month claimed 257 lives~\cite{Landslide_in_Ethiopia}.

To address the issue of landslides, a common approach involves the creation of landslide inventory maps, which store timestamp, location, and type of the event~\cite{guzzetti2012landslide}. 
In addition, to look ahead landslide events and provide early warning, landslide inventory maps can be built from various data sources, including aerial imagery from satellites~\cite{Li2016}, elevation models, and LiDAR altimetry~\cite{mckean2004objective}. 
Analyzing landslide by exploring normal images require high cost and skilled investigators who would verify images with tools and experience to delineate landslide boundaries~\cite{van2008spatial}. 
The output could be a finalized map where landslide areas are annotated with different colors, or it could undergo further analysis using Geographic Information Systems (GIS) software, facilitated by advancements in geospatial technology~\cite{guzzetti2012landslide}. 
Generally, this process, which heavily depends on human expertise, presents several challenges. 
Firstly, the low quality of given images in landslide-prone regions can compromise the accuracy of the mapping process, as it is time-consuming and labor-intensive. 
Secondly, experts often rely on specific `signatures', such as topological differences~\cite{GLENN2006131,KORUP2007578} or the principle that `the past and present are keys to the future'~\cite{SHU2019133557,PISANO20171147}, to identify landslide sites. 
The manual selection of these properties may lead to inconsistent results among researchers and could potentially overlook unknown factors contributing to future landslide occurrences.

Thanks to advancements in Earth Observation (EO) technologies, these techniques have significantly enhanced the accessibility and prevalence of remote sensing data from satellites such as Sentinel and Landsat~\cite{zhao2022overview}. 
These advancements include the introduction of multiple spectral bands capable of covering a wide range of visible wavelengths, with Sentinel-2 offering up to 13 bands, and the WorldView-3 satellite achieving a resolution as fine as 0.31 meters per pixel~\cite{vali2020deep}. 
Additionally, the rapid proliferation of unmanned aerial vehicles (UAVs)~\cite{alvarez2021uav} has established these technologies as mainstream tools in the study of landslide phenomena.
Furthermore, recent emergence of machine learning and deep learning has introduced innovative quantitative assessment methodologies for addressing the problem of landslide inventory mapping~\cite{mohan2021review}. 
Through iterative learning processes utilizing existing datasets, machine learning and deep learning models can uncover latent relationships between input data features and corresponding outcomes, providing implicit solutions for landslide segmentation without the necessity of designing complex mathematical models. 

\begin{table}[t]
    \caption{Bands' information in Landslide4Sense dataset~\cite{ls_data}} 
    \centering
    \scalebox{0.85}{
    \begin{tabular}{|c|c|c|l|} 
        \hline 
        \textbf{Band}& \textbf{Centre} & \textbf{Spatial} & \textbf{Description} \\
                     & \textbf{$\lambda$}(nm)  & \textbf{resolution}&                 \\
        \textbf{}& \textbf{} & \textbf{$\Delta$ $\lambda$}(m) & \textbf{} \\
        \hline
         Band 1 &  443 & 60 & Atmospheric correction (aerosol scattering)\\
        \hline
        Band 2 & 490 & 10 & Sensitive to vegetation senescing, carotenoid, \\
               &     &    & browning and soil background; atmospheric \\
               &     &    & correction (aerosol scattering)\\
        \hline
        Band 3 &  560 & 10 & Green peak, sensitive to total chlorophyll in\\
               &      &    &  vegetation\\
        \hline
        Band 4 &  665 & 10 & Maximum chlorophyll absorption\\
        \hline
        Band 5 &  705 & 20 & Position of red edge; consolidation of \\
               &      &    & atmospheric corrections/ fluorescence baseline.\\
               &      &    &\\
        \hline
        Band 6 &  740 & 20 & Position of red edge, atmospheric correction,\\
               &      &    &  retrieval of aerosol load.\\
        \hline
        Band 7 &  783 & 20 & Leaf Area Index (LAI)  \\
         \hline
        Band 8 &  842 & 10 & Edge of the Near-Infrared (NIR) plateau\\
        \hline
        Band 9 &  945 & 60 & Water vapour absorption, atmospheric \\
               &      &    & correction.\\
        \hline
        Band 10 &  1375 & 60 & Detection of thin cirrus for atmospheric \\
                &       &    &  correction. Sensitive to lignin, starch\\
        \hline
        Band 11 &  1610 & 20 &  and forest above ground biomass. Snow/ice\\
                &       &    &  /Scloud separation.\\
        \hline
        Band 12 &  1610 & 20 & Assessment of Mediterranean vegetation\\
                &       &    &  conditions. Distinction of clay soils for the\\
                &       &    &   monitoring of soil erosion. Distinction\\
                &       &    &   between live biomass, dead biomass and soil,  \\
                &       &    &  e.g. for burn scars mapping.\\
        \hline 
    \end{tabular}}
    \vspace{-0.2cm}
    \label{table:data1_info}
\end{table}

In recent years, research has often focused on traditional machine learning models such as Support Vector Machines (SVM)\cite{kavzoglu2019machine,rajmohan2021revamping}, Artificial Neural Networks (ANN)\cite{kavzoglu2019machine,park2013landslide}, and Random Forest (RF)~\cite{chen2017object,kavzoglu2019machine}. 
However, there is a notable shift in research trends toward deep learning models, such as U-Net~\cite{niu2022reg,chen2023landslide,chen2022drs}, DeepLab~\cite{du2021landslide}, and transformers~\cite{huang2023landslide,fu2021improved}, which are inspired by computer vision segmentation problems and often yield superior results. 
For instance, a comparative study on segmentation tasks in the Rasuwa district of the Himalayas demonstrated that the best machine learning model (a variant of Random Forest) achieved an F1 score of 82.07\% and a mean Intersection over Union (mIoU) score of 69.6\%, while the top-performing deep learning model attained an F1 score of 87.8\% and an mIoU score of 78.26\%~\cite{ghorbanzadeh2019evaluation}.
This indicates that deep-learning-based architectures perform more advances and prove potential for further improvement.
However, almost published papers for landslide detection or segmentation from remote sensing image, in which deep-learning-based models were proposed, used their own data from various resources or different settings~\cite{nodata_01, nodata_02, nodata_03, nodata_04, nodata_05, nodata_06}.
As these data were self-collected and has not published, this lead challenges to compare models' performance.
Additionally, authors focused on landslide events in certain regions which present different natural backgrounds.
These reasons causes challenges to evaluate cross-datasets which prevents the model development and evaluation.

Inspired by deep-learning approach and address the concerns mentioned above, we propose a deep-learning-based model for landslide detection and segmentation tasks by exploring remote sensing images in this paper. We mainly contribute:
\begin{itemize}
        \item We conducted extensive experiments to indicate the role of deep-learning techniques such as input features, network architectures, loss functions, etc. which are used to construct a deep-learning-based model for landslide segmentation. Given the experimental results, we proposed a novel network architecture, referred to as RMAU-NET, which achieved the high performance on both tasks of landslide detection and segmentation.
        \item To prove the proposed network architecture robust, we evaluate on three published and benchmark datasets of LandSlide4Sense~\cite{ls_data}, Bijie~\cite{ji2020landslide}, and Nepal~\cite{nepal_data}. The experimental results present our proposed model potential to integrate into real-life landslide observation systems.
\end{itemize}
            
The remainder of this paper is structured as follows: Section II describes remote sensing image (RSI) datasets proposed for landslide detection or segmentation. 
Given the RSI datasets, we define our tasks, suggest the evaluation metrics, data splitting, and present experimental settings in this section.
Section III presents our proposed baseline which is evaluated on the benchmark dataset of Landslide4Sense for the segmentation task.
Given the baseline in Section III, we evaluate a wide range of deep-learning techniques to further improve the baseline, achieving a novel network for both tasks of landslide detection and segmentation in Section IV.
Section IV evaluates our proposed network on different datasets of Landslide4Sense, Bijie, and Nepal, and then our proposed network is compared with the state-of-the-art systems.
Finally, Section V concludes the paper.
                
\begin{figure}[t]
    \centering
    \includegraphics[width=0.47\textwidth]{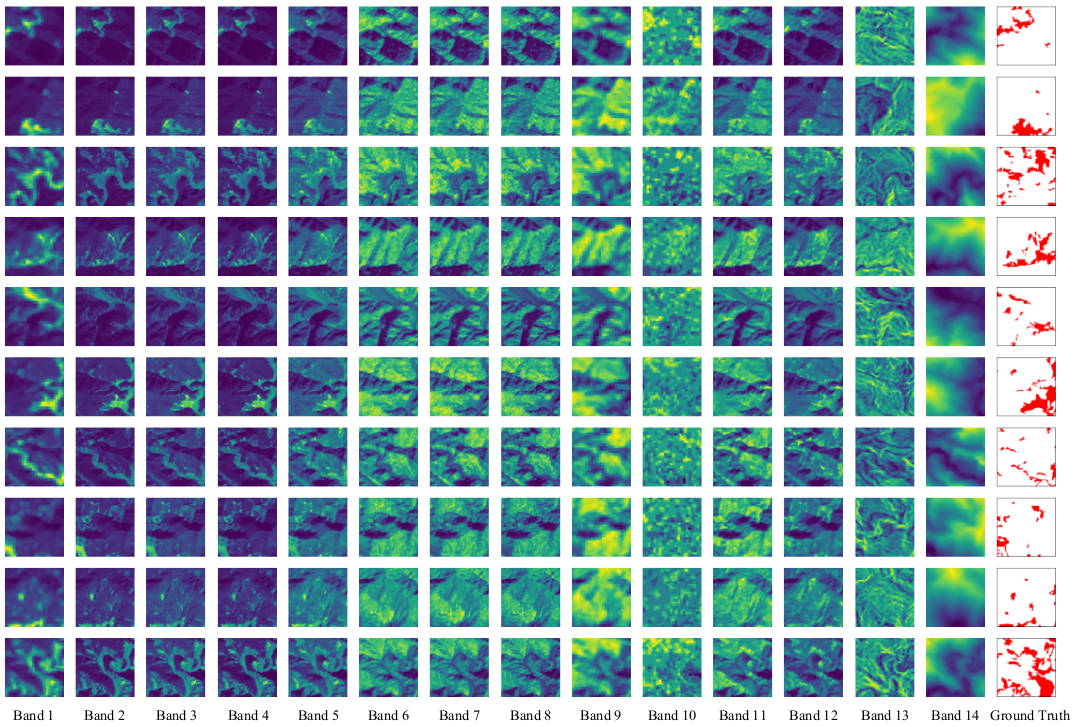}
    \caption{Some image samples with 14 band and corresponding mask from Landslide4Sense dataset~\cite{ls_data}}
    \label{fig:data1_sample}
\end{figure}
\begin{table}[t]
    \caption{Statistics of remote sensing images datasets for landslide detection or segmentation} 
    \centering
    \scalebox{0.7}{
    \begin{tabular}{|c|c|c|c|c|c|} 
        \hline 
        \textbf{Datasets}  &\textbf{Image}  &\textbf{Image}  &\textbf{Landslide}     &\textbf{Landslide}    &\textbf{Proposed} \\
                           &\textbf{Number} &\textbf{Size}                     & \textbf{Image Ratio}  &\textbf{Pixel Ratio}  &\textbf{Tasks}           \\
        
        \hline
        Landslide4Sense~\cite{ls_data} &3044 &128$\times$128$\times$14 &58\% &2.3\%  &Segmentation\\
        \hline   
        Bijie~\cite{ji2020landslide} &2773  &128$\times$128$\times$3 &27\% &3.9\% &Detection \\
        \hline
        Nepal~\cite{nepal_data}  &275  &128$\times$128 $\times$3 &- &0.8\%  &Segmentation \\ 
        \hline 
    \end{tabular}}
    \vspace{-0.2cm}
    \label{table:data_stat}
\end{table}

\section{Remote Sensing Image Datasets and Task Definition}
\label{dataset_define}
As the paper focuses on landslide detection and segmentation from remote sensing images, we first collect published and benchmark remote sensing image (RSI) datasets which involve in landslide events.
As our knowledge, there are three largest datasets proposed for landslide detection segmentation which have been published until now.
These are Landslide4Sense~\cite{ls_data}, Bijie~\cite{ji2020landslide}, and Nepal~\cite{nepal_data}.
Given these selected RSI datasets, we define two tasks of landslide detection and landslide segmentation, the evaluation metrics, dataset splitting, and experimental settings in this paper.

\subsection{Remote sensing image datasets}
\label{dataset}
\begin{figure}[t]
    \centering
    \includegraphics[width=0.47\textwidth]{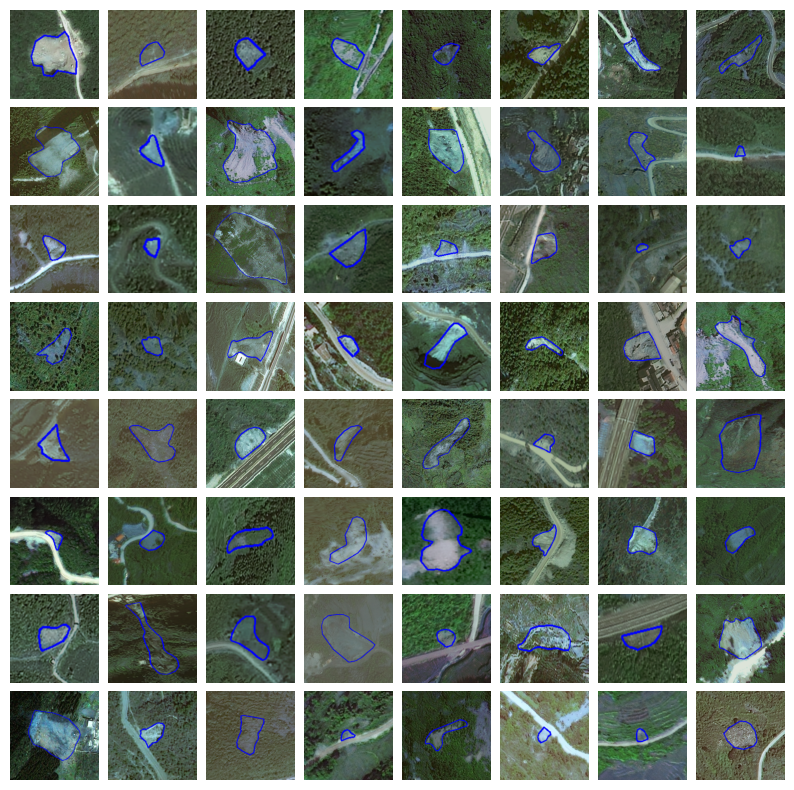}
    \caption{Some image samples from Bijie dataset~\cite{ji2020landslide}}
    \label{fig:Bijie_samples}
\end{figure}
\textbf{Landslide4Sense dataset~\cite{ls_data}} illustrates landslide appearances worldwide from 2015 to 2021. 
This dataset provides as an essential benchmark for analyzing landslide disaster in regions with heavy rainfall, earthquakes, or unstable geological conditions. 
The entire Landslide4Sense dataset comprises 3844 multi-spectral images each of which is a combination of 14 bands. 
In particular, bands from B1 to B12 (12 bands) are from the Sentinel-2 which are comprehensively described in Table~\ref{table:data1_info}.
B13 band is from slope data and B14 band was taken from the ALOS PALSAR satellite.
Each band presents the same size of 128$\times$128 and the mask data presents a binary image with the same dimensions of 128$\times$128. 
While white pixels in the mask indicate landslide-affected areas, black pixels indicate non-landslide areas. 
Additionally, each pixel in each layer corresponds to a real-world scale of 10 to 60 meters. 
Some image samples with 14 bands and the corresponding masks are comprehensively shown in Fig.~\ref{fig:data1_sample}
According to the statistics of Landslide4Sense dataset as shown in Table~\ref{table:data_stat}, the number of image contains landslide regions is approximately 58\% of all images in the dataset. 
However, the number of landslide region only take 2.3\% of all pixels in the dataset. 
This presents a significant imbalance between landslide and none-landsline regions which leads challenging to segmentation models.

\textbf{Bijie dataset~\cite{ji2020landslide}}: This dataset was collected from TripleSat Satellite for Bijie City, northwest of Guizhou Province, with an area of 26,853 \(\text{km}^2\) . 
The elevation of this region is from 457m to 2,900m combining with unstable geology and heavy rainfall, making it severe landslide location in China. 
After collecting from the satellite, the images are transformed into RGB format which presets the size of 128$\times$128$\times$3.
Fig.~\ref{fig:Bijie_samples} illustrates some RGB image samples from the dataset (i.e., The blue lines in each image mark landslide regions). 
In total, the Bijie dataset comprises 770 landslide images. 
Additionally, the dataset includes 2,003 images of none-landslide regions in Bijie city. 
This indicates an imbalance between landslide and none-landslide image samples.
Regarding landslide and none-landslide pixels, the statistics in Table~\ref{table:data_stat} presents a ratio of 3.9/96.1 which shows a significant imbalance and same as Landslide4Sense dataset.

\begin{figure}[t]
    \centering
    \includegraphics[width=0.4\textwidth]{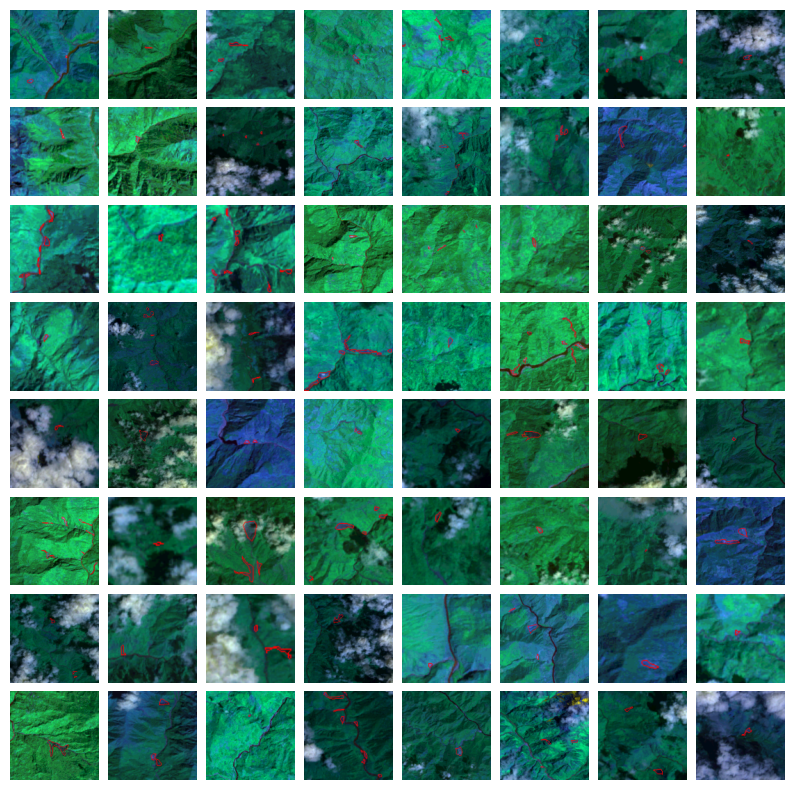}
    \caption{Some image samples from Nepal Dataset~\cite{nepal_data}}
    \label{fig:Nepal_samples}
\end{figure}
    
\textbf{Nepal~\cite{nepal_data}}: To collect the data, geologists first extract locations and time of landslide events.
Given these information, remote sensing images from Landsat-8 satellite were obtained and manually verified.
Same as Bijie dataset, the collected images from the satellite were transformed into RBG format with the size of 128$\times$128$\times$3.
Some image samples are shown in Fig.~\ref{fig:Nepal_samples}.
Totally, Nepal dataset~\cite{nepal_data} comprises 230 images with landslide events. 
According to statistics in Table~\ref{table:data_stat}, the total number of landslide pixels is very small, only 0.8\% of the entire dataset which present a significant imbalance regarding pixel level. 

\begin{figure*}[ht]
    \centering
    \includegraphics[width=\textwidth]{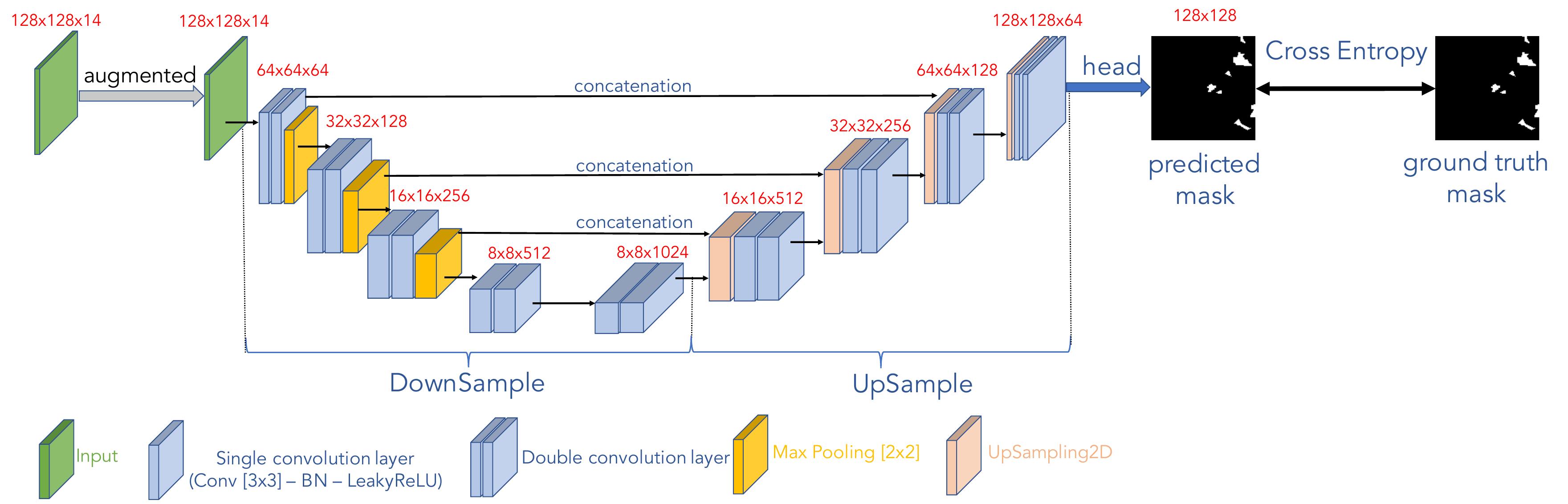}
    \caption{The proposed U-Net baseline architecture for the landslide segmentation.}
    \label{f1_baseline}
\end{figure*}

\subsection{Task definition}
\label{task_def}

Given the published remote sensing image datasets which involve in landslide events, it reveals that each dataset was proposed for either landslide detection or landslide segmentation.
Indeed, published papers using the Nepal dataset have omitted the segmentation task~\cite{nepal_data}, \cite{chen2018practical}, \cite{yu2020landslide} as this dataset only provides remote sensing images with landslide regions. 
In contrast, the Bijie dataset includes a significant number of non-landslide images, which led to the primary focus on classification task in relevant papers of~\cite{ji2020landslide}, \cite{qin2021landslide}, \cite{li2024landslide}. 
Regarding LandSlide4Sense dataset, it was proposed for a challenge and the competition metrics are for the landslide segmentation task~\cite{Ghorbanzadeh_2022}, \cite{ls_data}.
This inspires us to evaluate our proposed model on both tasks of landslide detection and segmentation on Bijie and Landslide4Sense datasets, and only segmentation task on then Nepal dataset.

Compare among these datasets, it can be seen that Landslide4Sense presents the largest image number with a balance landslide/none-landslide image.
The dataset also show diverse landslide region collected from various places over the world.
We, therefore, select Landslide4Sense dataset to evaluate our proposed baseline.
We then further improve the baseline by applying and evaluating various deep-learning techniques such as input features, network architecture, loss function, post-processing, etc.
Given evaluation results, we propose the best model configuration for both tasks of landslide detection and segmentation.
We then evaluate the best model on the remaining datasets of Bijie, Nepal, and compare with the state-of-the-art systems.

\subsection{Dataset splitting}
As the LandSlide4Sense dataset~\cite{ls_data} has not published the label of test set (800 images). 
We split the training set (3044 images) into two parts, using an 80:20 ratio, for training and testing, respectively. 
For the Bijie dataset~\cite{ji2020landslide}, we omit the suggestion in the paper \cite{qin2021landslide}, remaining the 70:30 ratio for the train and test sets.
Regarding Nepal dataset~\cite{nepal_data}, it was suggested to divide the entire dataset into training, validation, and test sets. 
Hence, we omit this suggestion and keep the training and validation sets for model development while the test set is used for the evaluation.

\subsection{Evaluation metrics}
To evaluate our proposed model, we report the F1-score, Precision, Recall, and the mean Intersection over Union (mIoU) on the pixel level for the segmentation task. 
Regarding the detection task, F1-score, Precision, and Recall on the image level are provided.

\subsection{Experimental settings}
We construct our proposed deep neural networks with Tensorflow framework. 
All deep neural networks are trained for 30 epochs on Titan RTX 24GB gpu. 
All evaluating models in this paper use Adam~\cite{Adam} for the optimization in the training process.
      
\section{The proposed baseline model}
\label{baseline}

As mentioned in Section~\ref{task_def}, we first propose a baseline model for landslide segmentation and evaluate multiple deep-learning techniques with the baseline on LandSlide4Sense dataset.
As shown in Fig.~\ref{f1_baseline}, the baseline consists of three main components: The online data augmentation, the U-Net backbone network architecture, and the head with the loss function for the segmentation task.

\subsection{The online data augmentation}
Given input remote sensing images, we first apply two data augmentation methods of rotation and cutmix.
In particular, each image is randomly rotated using angles of 90, 180, or 270 degrees to generate a new images, referred to as the rotation augmentation.
Then, landslide regions from random landslide images are cut and mixed with the current processing image, referred to as the cutmix augmentation~\cite{cutmix_image}. 
As these two data augmentation methods are applied on batch of remote sensing images during the training process, we referred to as the online data augmentation.

\subsection{The U-Net based backbone architecture}
As Fig.~\ref{f1_baseline} shows, the U-Net backbone consists of the downsample and upsample networks. 
Both downsample and upsample networks leverage the double convolutional architectures.
Each of double convolutional architecture presents two single convolutional layers performing a convolutional layer (Conv),  batchnorm layer (BN)~\cite{batchnorm}, and Leaky Rectifine Linear Unit (LeakyReLU)~\cite{leak_relu} in the order.
For down-sampling, MaxPooling layer is applied.
Meanwhile, UpSampling2D layer is used for up-sampling.

\subsection{The head and loss function}
Given the output feature map of 128$\times$128$\times$64 from the U-Net backbone, a global average pooling layer across the channel dimension is applied to obtain the predicted mask of 128$\times$128. 
The predicted mask is then compared with the ground truth mask using Cross-Entropy loss function 

\begin{equation}
    \label{eq:loss_func}
    Loss_{EN}(\theta) = -\frac{1}{N}\sum_{i=1}^{N}y_i .log \left\{\hat{y}_{i}(\theta) \right\} + \frac{\lambda}{2}.||\theta||_{2}^{2}
\end{equation}
where \(Loss(\theta)\) is the loss function over all trainable parameters \(\theta\), constant \(\lambda\) is set to $0.0001$,  $y_{i}$ and $\hat{y}_{i}$  are expected and predicted $N=16384$ pixels from the feature map of 128$\times$128.

\subsection{Evaluate the proposed baseline on Landslide4Sense dataset} 
We evaluate the proposed baseline on Landslide4Sense dataset with the segmentation task.
To this end, we first train the proposed baseline.
After the training process, we feed the test images of 128$\times$128$\times$14 into the baseline, obtain the predicted mask.
The predicted mask is then compared with the ground truth, compute the F1 and mIoU scores on the pixel level.
As Table~\ref{table:res_t1} shows, the baseline achieves F1 and mIoU scores of 67.83 and 60.01, respectively. 

\begin{table}[t]
    \caption{Evaluate the proposed baseline on Landslide4Sense dataset with the segmentation task} 
    \vspace{-0.2cm}
    \centering
    \scalebox{1.0}{
    \begin{tabular}{|l|c|c|} 
        \hline 
        \textbf{Network}   &  \textbf{F1 score}  &  \textbf{mIoU} \\       
        \hline
        U-Net baseline & 67.83 & 60.01\\
       \hline 
    \end{tabular}
    }
    \vspace{-0.3cm}
    \label{table:res_t1} 
\end{table}
\begin{table}[t]
    \caption{Evaluate the effect of the loss functions.} 
    \vspace{-0.2cm}
    \centering
    \scalebox{1.0}{
    \begin{tabular}{|l|c|c|} 
        \hline 
        \textbf{Networks \& Loss}   &  \textbf{F1 score}  &  \textbf{mIoU} \\       
        \hline
         U-Net  \& Cross Entropy (baseline)   & 67.83 & 60.01\\
         U-Net  \&  Focal Loss      & \textbf{68.28}  &\textbf{60.37} \\
         U-Net  \&  Log-Cosh Loss   &67.73  &59.95  \\
         U-Net  \&  IoU Loss        &\textbf{68.20}  &\textbf{60.23}  \\
         U-Net  \&  Tversky Loss    &66.21  &58.54  \\
         U-Net  \&  Lovasz Loss     &64.61  &57.37  \\
         U-Net  \&  Boundary Loss   &60.14  &54.92  \\
         U-Net  \&  Center Loss     &65.61  &58.21  \\    
         \hline                                      
         U-Net  \&  Focal \& IoU Loss      &\textbf{69.05} &\textbf{61.14} \\
       \hline 
    \end{tabular}
    }
    \vspace{-0.3cm}
    \label{table:res_t1} 
\end{table}
\begin{table}[t]
    \caption{New band data is generated from 14 original bands in Landslide4Sense dataset.} 
    \vspace{-0.2cm}
    \centering
    \scalebox{1.0}{
    \begin{tabular}{|l|l|} 
        \hline 
        \textbf{New band data}   &  \textbf{Formula / Method} \\
        \hline
        Band 15 to Band 17 &  $(x - x\_min)/(x\_max - x\_min)$ \\
        Band 18: NDVI & $(B8-B4)/(B8+B4)$\\ 
        Band 19: NDMI &  $(B8-B11)/(B8+B11)$\\ 
        Band 20: NBR &  $(B8-B12)/(B8+B12)$ \\ 
        Band 21:Gray &  $(B2+B3+B4)/3$ \\ 
        Band 22 to Band 23 & Gausian and Median filters\\
        Band 24 to Band 25 & Image gradients across length and width\\
        Band 26 & Canny Edge detector\\
        \hline 
    \end{tabular}
    }
    \vspace{-0.2cm}
    \label{table:band} 
\end{table}
\begin{table}[t]
    \caption{Evaluate the effect of the input feature (U-Net*: U-Net baseline with the combined loss function).} 
    \vspace{-0.2cm}
    \centering
    \scalebox{1.0}{
    \begin{tabular}{|l|c|c|} 
        \hline 
        \textbf{Network \& band data}   &  \textbf{F1 score}  &  \textbf{mIoU} \\
        \hline
         U-Net* \& Original 14 bands                     &69.05 &61.14\\
         U-Net* \& Original 14 bands \& bands 15 to 17   &69.39 &61.22\\
         U-Net* \& Original 14 bands \& bands 15 to 21   &69.83 &60.97\\
         U-Net* \& Original 14 bands \& bands 15 to 23   &\textbf{69.96} &\textbf{61.76}\\
         U-Net* \& Original 14 bands \& bands 15 to 25   &69.91 &61.64\\
         U-Net* \& Original 14 bands \& bands 15 to 26   &68.54 &60.65\\
       \hline 
    \end{tabular}
    }
    \label{table:res_t2} 
\end{table}
\begin{figure}[t]
    \centering
    \includegraphics[width=0.4\textwidth]{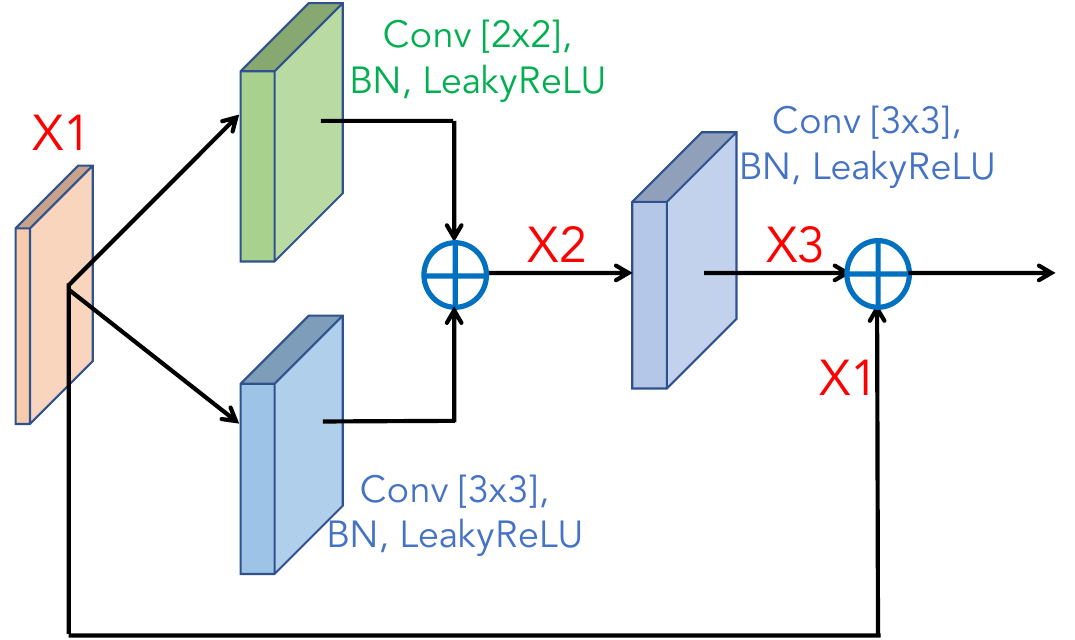}
    \caption{The proposed Residual-Convolutional layer.}
    \label{f13_res_lay}
\end{figure}

\begin{table}[t]
    \caption{Evaluate the effect of the multiple heads (U-Net+: U-Net baseline with 23 band data and combined loss function).} 
    \vspace{-0.2cm}
    \centering
    \scalebox{1.0}{
    \begin{tabular}{|l|c|c|} 
        \hline 
        \textbf{Network}   &  \textbf{F1 score}  &  \textbf{mIoU} \\
        \hline
         U-Net+                     &69.96 &61.76\\
         U-Net+ \& Multiple heads   &\textbf{70.45} &\textbf{62.19}\\
         
       \hline 
    \end{tabular}
    }
    \label{table:res_t3} 
\end{table}

\begin{figure}[t]
    \centering
    \includegraphics[width=0.5\textwidth]{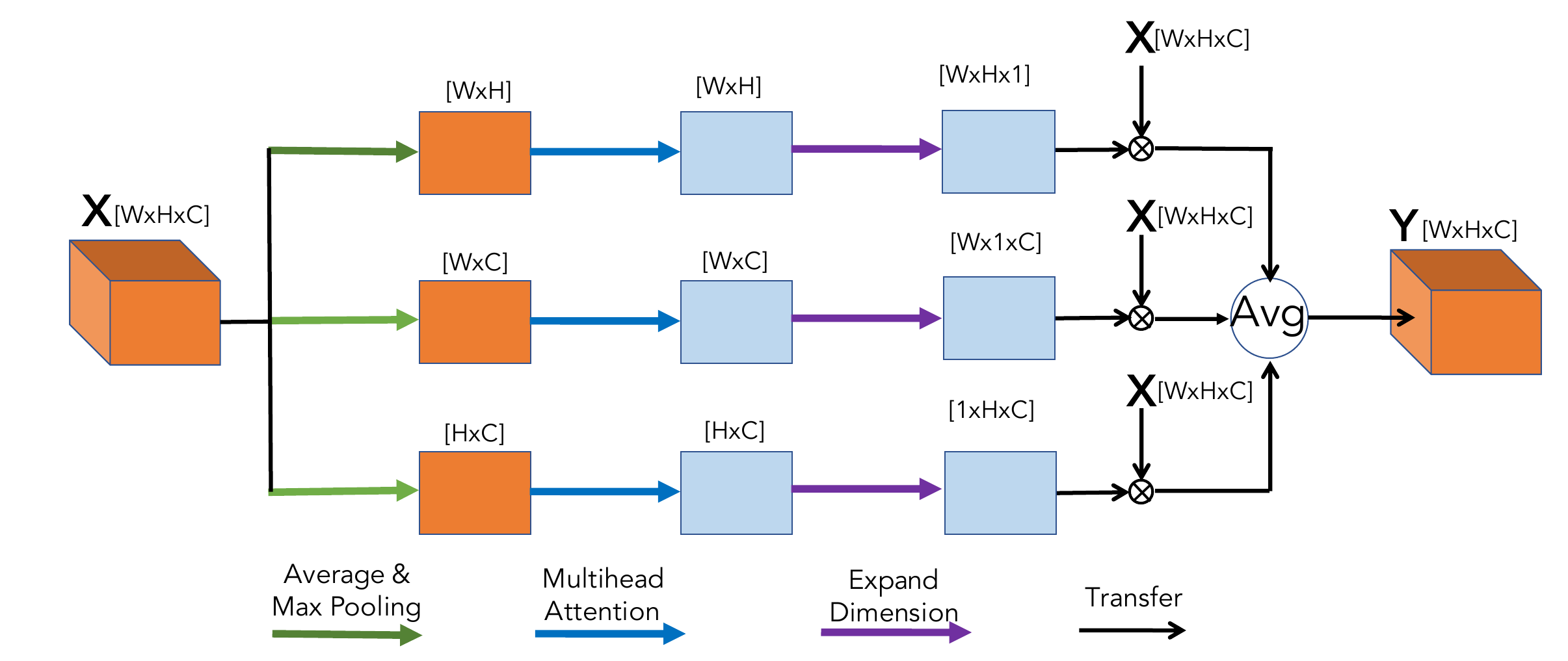}
    \caption{The proposed multihead-attention layer.}
    \label{f14_multihead}
\end{figure}
\begin{table}[t]
    \caption{Evaluate the network architecture improvements (U-Net\ding{61}: U-Net baseline with 23 band data, combined loss function, and multiple resolution heads).} 
    \vspace{-0.2cm}
    \centering
    \scalebox{1.0}{
    \begin{tabular}{|l|c|c|} 
        \hline 
        \textbf{Networks}   &  \textbf{F1 score}  &  \textbf{mIoU} \\
        \hline
         U-Net\ding{61}                       &70.45 &62.19\\
         DeepLab-V3                   &68.30 &60.49 \\ 
         MobileNet-V3                 &60.26 &54.72 \\ 
         EfficientNet-V2              &64.51 &57.27 \\ 
         \hline
         U-Net\ding{61}  \& CBAM Att.         &70.82 &62.53 \\                
         U-Net\ding{61}  \& SE Att.           &71.26 &62.86 \\
         U-Net\ding{61}  \& Multihead Att.    &71.45 &63.05 \\
         U-Net\ding{61}  \& Res-Conv          &72.07 &63.45\\
         \hline
         U-Net\ding{61}  \& Multihead Att. \& Res-Conv &\textbf{72.42} &\textbf{63.88} \\
       \hline 
    \end{tabular}
    }
    \label{table:res_t4} 
\end{table}
\begin{table}[t]
    \caption{Evaluate the threshold values} 
    \vspace{-0.2cm}
    \centering
    \scalebox{1.0}{
    \begin{tabular}{|c|c|c|} 
        \hline 
        \textbf{Threshold Values}   &  \textbf{F1 score}  &  \textbf{mIoU} \\
        \hline
        None  &72.42 &63.88 \\
        0.4   &71.78 &63.29 \\
        0.5   &72.42 &63.88 \\
        0.6   &72.60 &64.02 \\
        0.75  &73.02 &64.43 \\
        0.85  &73.13 &64.69 \\
        0.9   &73.68 &65.07 \\
        \textbf{0.95}  &\textbf{74.63} &\textbf{65.97} \\
        0.99  &73.11 &64.68 \\      
       \hline 
    \end{tabular}
    }
    \label{table:res_t5} 
\end{table}
\section{Evaluate deep learning techniques to improve the baseline}
Given the baseline, we now conduct extensive experiments applying a wide range of deep-learning techniques of loss function, input features, network architectures, optimization algorithms, and post processing methods in the order.
We evaluate whether these techniques are effective to further improve the baseline performance.

\begin{figure*}[t]
    \centering
    \includegraphics[width=1.0\textwidth]{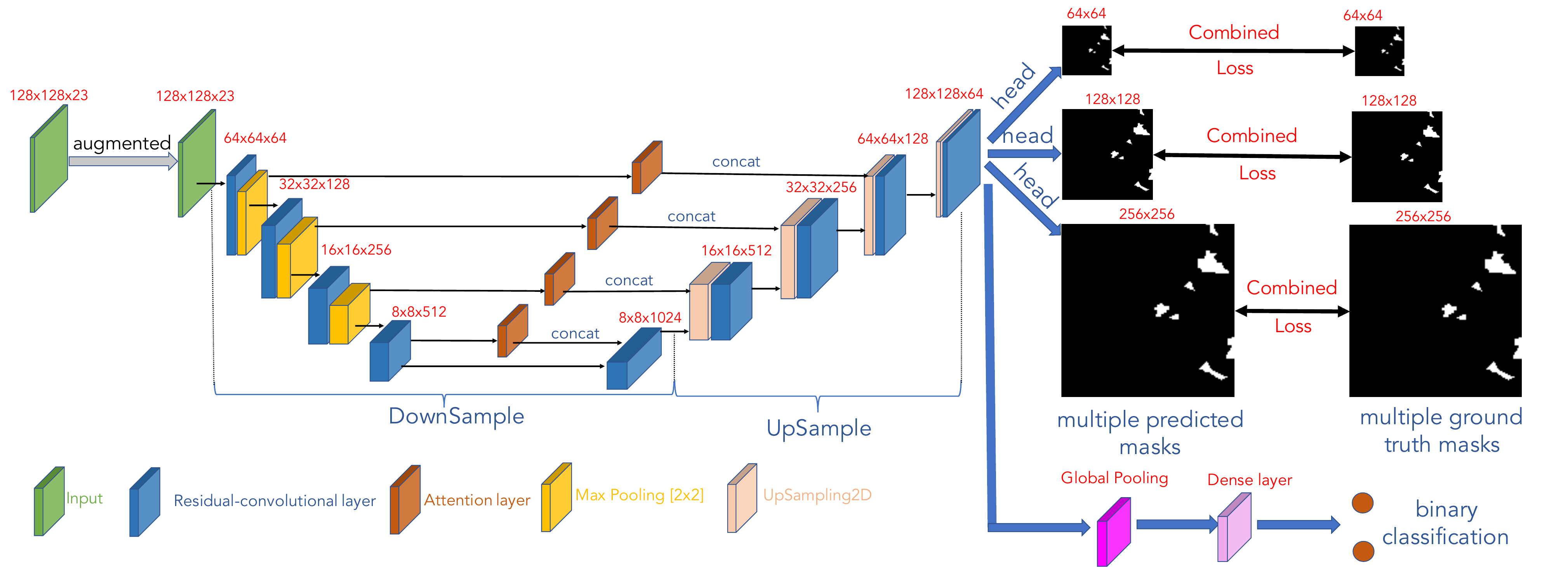}
    \caption{The proposed RMAU-NET architecture.}
    \label{f1_final_model}
\end{figure*}
\subsection{Evaluate loss functions}
\label{loss}

Among deep-learning techniques mentioned above, we first evaluate the role of loss functions.
We tackle the imbalance issue between landslide pixels and non-landslide pixels in the segmentation task by evaluating a wide range of loss functions of Focal loss~\cite{focal_loss}, Log-Cosh loss~\cite{logcosh_loss}, IoU loss~\cite{iou_loss}, Tversky loss~\cite{tversky_loss}, Lovasz loss~\cite{lovasz_loss}, Boundary loss~\cite{bou_loss}, and Center loss~\cite{center_loss}. 
While the online data augmentation and U-Net backbone in the proposed baseline are remained, the Cross-Entropy loss in the head is replaced by the loss functions in Table~\ref{table:res_t1} to perform the segmentation task on Landslide4Sense dataset.

As the experimental results shown in Table~\ref{table:res_t2}, Focal loss and IoU loss outperform the other loss functions.
This inspires us to combine both Focal loss and IoU loss.
The combination of Focal loss and IoU loss is defined by 
\begin{eqnarray}
    \label{eq:lin_com}
     Loss =  \alpha.Loss_{Focal} + (1-\alpha).Loss_{IoU},
\end{eqnarray}
where $\alpha$ presents the portion between two loss functions, which is empirically set to 0.5.
Compare to the baseline with Cross-Entropy loss, the combination of Focal loss and IoU loss yields improvements of 1.22 regarding the F1-score and 1.13 for the mIoU score.

\subsection{Evaluate input feature}
\label{feature}
Inspired by~\cite{inspire_01}, we evaluate whether generating band data is effective to enrich the features of remote sensing images, then improving the segmentation performance.
Therefore, we generate 12 new band data from the 14 original bands of a remote sensing image in Landslide4Sense dataset. The methods used to generate new band data are presented in Table~\ref{table:band} and summarized by:

\begin{itemize}
\item Bands 15 to 17 are generated by applying RGB normalization on bands B2, B3 and B4.
\item Bands 18 to 21 represent remote sensing indices (NDVI, NDMI, NBR) and a grayscale image.
\item Bands 22 and 23 are generated by applying Gaussian and median filters with kernel size of [10$\times$10].
\item Bands 24 and 25 are calculated from the image gradient (across length and width dimensions).
\item Band 26 presents the result of using Canny edge detector.
\end{itemize}

As the experimental results are shown in Table~\ref{table:res_t2}, it indicates that adding generating band data is effective to improve the performance of the segmentation task. 
Particularly, adding bands from 15 to 23 presents the most significant improvement of 0.91 and 0.62 for F1 score and mIoU score, respectively.

\subsection{Evaluate the network architectures}
\label{architecture}

Regarding the network architecture, we propose two main improvements: (1) The proposed multiple head resolution instead of the single head architecture in the baseline; (2) The combination of residual-convolutional and attention layers to replace the traditional convolutional layer in U-Net backbone.
The detailed improvements are comprehensive described in the Fig.~\ref{f1_baseline}. 

For the first improvement,  we are inspired by applying an ensemble of multiple predicted masks with different resolutions to enhance the system performance.
In particular, instead of using only one head block to generate one predicted mask of 128$\times$128, we add two more head blocks to generate two other predicted masks: 256$\times$256 and 64$\times$64.
As a result, the final predicted result is obtained from an average of three predicted output masks. 
The experimental result on Landslide4Sense dataset shown in Table~\ref{table:res_t3} indicates that applying multiple heads helps to enhance the segmentation performance, further improving 0.49 and 0.53 on F1 score and mIoU score, respectively.
Notably, this experiment remains the advances of 23 band data and the combination of Focal and IoU loss functions in previous experiments.


Regarding the U-Net backbone, we first evaluate whether U-Net based architecture is the most effective model for the segmentation task.
In particular, we replace the U-Net backbone by Deeplab-V3~\cite{deepv3}, MobileNet-V3~\cite{mobilenetv3} and EfficientNet-V2~\cite{effnetv2} architectures.
Secondly, we are inspired that the multiple kernel sizes and a residual based architecture is more effective to capture distinct features of feature maps rather than a conventional convolutional layer.  
We therefore develop an architecture of a residual-convolutional layer (Res-Conv) as shown in Fig.~\ref{f13_res_lay} which is used to replace the double convolution layer in both the downsample and upsample architecture in the baseline.


We further improve the network architecture by applying an attention layer after every convolutional layer in both the downsample and upsample architectures.
The attention weights generated by the proposed attention layer effectively enforces the neural network to focus on landslide regions on the feature maps inside the network.  
We evaluate three types of attention schemes: SE~\cite{se_att} attention, CBAM~\cite{cbam_att} attention, and our proposed multi-head attention in our published paper~\cite{our_att}.
Rather only focusing on certain dimension of a feature map, our proposed multi-head attention as shown in Fig.~\ref{f14_multihead} explores all three dimensions of the feature map.
In particular, given an input feature map $\mathbf{X}$ with a size of [W$\times$H$\times$C] where W, H, and C presents width, height, and channel dimensions, the feature map $\mathbf{X}$ is reduced size across three dimensions using both max and average pooling layers, generating new two-dimensional feature maps.
Then, the traditional multi-head attention~\cite{multihead_att} is applied to each two-dimensional feature maps before multiplying with the original three-dimensional feature map $\mathbf{X}$.

As the experimental results are shown in Table~\ref{table:res_t4}, U-Net based architecture is more effective than evaluating network architectures of DeepLab-V3, MobileNet-V3, EfficientNet-V2.
Regarding applying attention and Res-Conv layer, both techniques help to improve the segmentation performance.
When these techniques are combined, it helps to improve F1 score and mIoU score by 1.97 and 1.69 respectively.
Notable, this experiment reuses the advances of previous experimental results with using 23 band data, combined loss function, and multiple resolution heads




\subsection{Evaluate post processing}
Given the advances of using 23 band data, combined loss function, multiple resolution heads, and combination of multihead attention and Res-Conv layer regarding U-Net backbone, we finally evaluate the role of post processing step.
In particular, we apply a threshold value to decide whether a pixel is referred to as the landslide or none-landslide.
The Table~\ref{table:res_t5} indicates that we obtained the best F1 score of 74.463 and mIoU score of 65.97 at the threshold value of 0.95.

\section{Propose RMAU-NET for landslide detection and segmentation}
As the extensive experiments were conducted above, we indicate that leveraging multiple deep-learning techniques of combined loss (IoU loss and Focal loss), 23 band data (8 generating band data and 14 original band data), multiple resolution heads, a combination of Res-Conv layer and our proposed multihead attention layer, and post-processing with certain threshold shows effective to further improve the segmentation performance.
The Table~\ref{table:res_t6} comprehensively describes the improvement from each technique with the significant enhancement from network architecture improvement (applying Res-Conv layer and Multihead attention layer) and threshold-based post-processing.

Given these advanced techniques, we propose a novel network architecture for both tasks of landslide segmentation and detection, referred to as RMAU-NET.
As the novel network is shown in Fig.~\ref{f1_final_model}, all advanced deep learning techniques, which are evaluated in the previous section, are applied.
To adapt the landslide detection task, the global pooling layer is applied on the feature map of 128$\times$128$\times$64 before flattening, and going through a dense layer for the binary classification of landslide and none-landslide.
The proposed RMAU-NET is then evaluated with other datasets of Bijie~\cite{ji2020landslide}, and Nepal~\cite{nepal_data}, achieving the state-of-the-art results on both landslide detection and segmentation tasks as shown in Table~\ref{table:res_t7}.
Some segmentation results obtained from RMAU-NET model on LandSlide4Sense dataset are also shown in Fig.~\ref{f1_final_model}

\begin{table*}[t]
    \caption{Apply deep learning techniques to further imrpove the U-Net baseline for landslide segmentation on Landslide4Sense dataset} 
    \vspace{-0.2cm}
    \centering
    \scalebox{1.0}{
    \begin{tabular}{|c|c|c|c|c|c|c|c|c|} 
        \hline 
        \textbf{Network}   &\textbf{Combined loss}  &\textbf{23 band data}  &\textbf{Multiple heads} &\textbf{Res-Conv \&} &\textbf{Threshold} &  \textbf{F1 score}  &  \textbf{mIoU} \\
        &&&&\textbf{Multihead Att.} & &&\\
        \hline
        U-Net w/ &- &-&-&-&-  &67.83 &60.01 \\
        U-Net w/ &\checkmark &-&-&-&-  &69.05 &61.14 \\
        U-Net w/ &\checkmark &\checkmark &-&-&- &69.96 &61.76 \\
        U-Net w/ &\checkmark &\checkmark &\checkmark &-&- &70.45 &62.19 \\        
        U-Net w/ &\checkmark &\checkmark &\checkmark &\checkmark &- &72.42 &63.88 \\
        U-Net w/ &\checkmark &\checkmark &\checkmark &\checkmark &\checkmark  &74.63 &65.97 \\
       \hline 
    \end{tabular}
    }
    \label{table:res_t6} 
\end{table*}

\begin{table*}[t]
\centering
\caption{Performance Comparison on LandSlide4Sense, Nepal, and Bijie datasets}
\begin{tabular}{|c|c|cccc|cccc|}
   \hline
   \multirow{2}{*}{ \textbf{Dataset}} & \multirow{2}{*}{\textbf{Model}} &\multicolumn{4}{c|}{\textbf{Segmentation Task}} & \multicolumn{4}{c|}{\textbf{Detection Task}} \\
  \cline{3-10} 
   & &F1 score & Precision & Recall &mIoU & F1 Score & Accuracy & Precision & Recall \\
   \hline 
    Landslide4Sense         &Our proposed RMAU-NET &76.90 &72.11 &82.38 &65.97 &98.23 &97.10 &98.58 &97.89\\    
   \hline
   \multirow{4}{*}{Nepal}   &ResNet~\cite{nepal_data} &60.00 &55.00 &65.00 &n/a &n/a &n/a &n/a &n/a\\
         &U-Net~\cite{nepal_data} &67.00 &61.00 &74.00 &n/a &n/a &n/a &n/a &n/a\\  
         &Our proposed RMAU-NET &69.43 &62.12 &78.79 &76.88 &n/a &n/a &n/a &n/a\\  
   \hline  
   \multirow{4}{*}{Bijie}      &DDTL~\cite{ji2020landslide}          &n/a &n/a &n/a &n/a &n/a &79.69 &n/a &n/a\\  
                               &DDTL+SE~\cite{ji2020landslide}       &n/a &n/a &n/a &n/a &n/a &93.36 &n/a &n/a\\  
                               &DDTL+CBAM~\cite{ji2020landslide}     &n/a &n/a &n/a &n/a &n/a &95.89 &n/a &n/a\\  
                               &Improved DDTL~\cite{ji2020landslide} &n/a &n/a &n/a &n/a &n/a &96.03 &n/a &n/a\\  
                               &Our proposed RMAU-NET &74.34 &74.54 &74.14 &57.33  &93.83 &96.63 &95.52 &92.21\\     
\hline
\end{tabular}
\label{table:res_t7} 
\end{table*}


\begin{figure*}[t]
    \centering
    \includegraphics[width=1.0\textwidth]{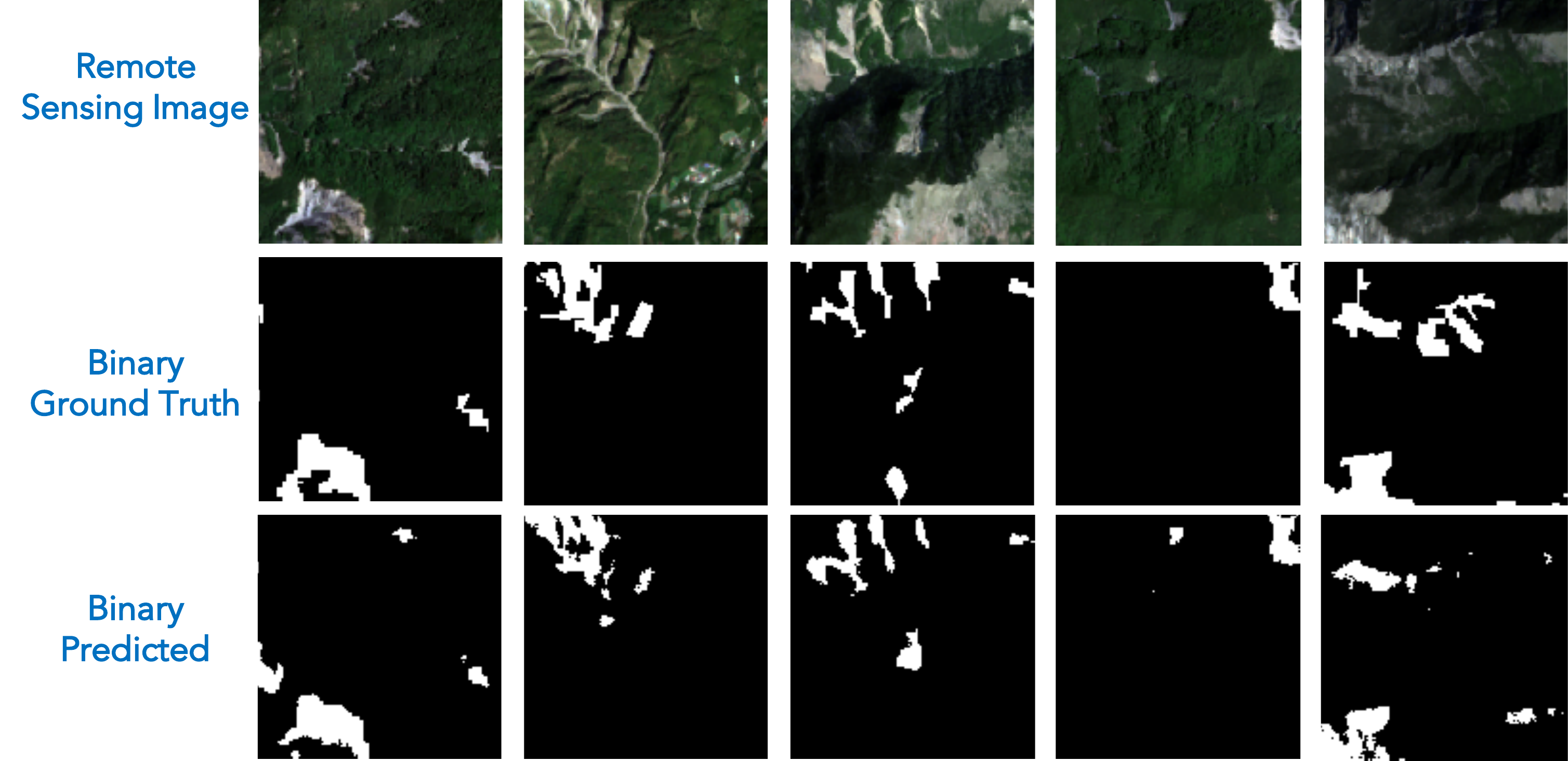}
    \caption{Segmentation results obtained from RMAU-NET model on LandSlide4Sense remote sensing images.}
    \label{f1_final_model}
\end{figure*}

\section{Conclusion}

We have presented a deep-learning-based approach for landslide detection and segmentation from remote sensing imagery.
By evaluating the effects of various improvements of feature engineering, network architecture, loss functions, optimization algorithms, and post processing, we finally construct the RMAU-NET which bases on U-Net architecture. 
The extensive experiments prove that our proposed RMAU-NET is robust on different benchmark datasets of Landslide4Sense, Bijie, and Nepan which shows potential to apply for remote-sensing-image-based landslide analysis systems.




\bibliographystyle{IEEEbib}
\bibliography{refs}
\end{document}